RESEARCH

# MediHive: A Decentralized Agent Collective for Medical Reasoning

**Xiaoyang Wang[1] · Christopher C. Yang[1]**

## Abstract

Large language models (LLMs) have revolutionized medical reasoning tasks, yet single-agent systems often falter on complex, interdisciplinary problems requiring robust handling of uncertainty and conflicting evidence. Multi-agent systems (MAS) leveraging LLMs enable collaborative intelligence, but prevailing centralized architectures suffer from scalability bottlenecks, single points of failure, and role confusion in resource-constrained environments. Decentralized MAS (D-MAS) promise enhanced autonomy and resilience via peer-to-peer interactions, but their application to high-stakes healthcare domains remains underexplored. We introduce MediHive, a novel decentralized multi-agent framework for medical question answering that integrates a shared memory pool with iterative fusion mechanisms. MediHive deploys LLM-based agents that autonomously self-assign specialized roles, conduct initial analyses, detect divergences through conditional evidence-based debates, and locally fuse peer insights over multiple rounds to achieve consensus. Empirically, MediHive outperforms single-LLM and centralized baselines on MedQA and PubMedQA datasets, attaining accuracies of 84.3% and 78.4%, respectively. Our work advances scalable, fault-tolerant D-MAS for medical AI, addressing key limitations of centralized designs while demonstrating superior performance in reasoning-intensive tasks.

✉ Xiaoyang Wang
xw388@drexel.edu

✉ Christopher C. Yang
chris.yang@drexel.edu

[1] College of Computing and Informatics, Drexel University, Philadelphia, PA 19104, USA

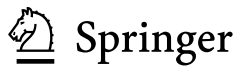

## 1 Introduction

The advent of Large Language Models (LLMs) has transformed medical reasoning, enabling advanced capabilities in diagnostic support, treatment planning, and knowledge synthesis across healthcare domains such as disease diagnosis [1], patient management, and clinical decision making [2–4]. However, single LLM-based agents frequently encounter limitations in handling the intricate, multifaceted nature of medical problems, which demand interdisciplinary expertise, real-time adaptation to patient-specific data, and robust handling of uncertainties like incomplete information or conflicting evidence. This has driven the emergence of multi-agent systems (MAS) in healthcare, where autonomous agents collaborate to simulate multi-disciplinary team (MDT) discussions, decompose diagnostic tasks, share insights, and refine reasoning through iterative interactions, thereby achieving collective intelligence tailored to medical contexts [5].

In LLM-based MAS for medical reasoning, agents harness the natural language processing, logical inference, and knowledge integration strengths of LLMs to interpret patient data [6], query medical databases, and generate evidence-based recommendations collaboratively. Conventional MAS architectures in medicine often rely on centralized coordination, such as a primary agent overseeing workflows or static hierarchies for task allocation [7]. While effective in controlled environments, these centralized approaches suffer from several drawbacks. First, they exhibit limited scalability, as the central agent becomes a bottleneck when handling large numbers of agents or high-volume queries, leading to performance degradation in expansive systems. Second, they introduce a single point of failure, where the malfunction or overload of the coordinator can halt the entire system, compromising reliability in critical applications. Third, in resource-constrained scenarios where a single LLM instance simulates multiple positions within the MAS, there is a heightened risk of confusion, leakage, or disarray among roles and knowledge bases. This can manifest as agents inadvertently blending personas, leading to inconsistent reasoning or diluted expertise, ultimately undermining the system's integrity and output quality [8].

Decentralized Multi-Agent Systems (D-MAS) emerge as a promising alternative, addressing these limitations by empowering agents to operate autonomously through peer-to-peer communication and self-organization, without reliance on a central authority. This approach enhances scalability by distributing computational load and decision-making, improves fault tolerance as the system can persist even if individual agents fail, and mitigates role confusion by allowing each agent—potentially instantiated as separate LLM calls—to maintain isolated yet collaborative knowledge states. Existing D-MAS works, such as AgentNet's self-evolving roles and decentralized collaboration [9], or Symphony's privacy-saving orchestration with low overhead [10], have demonstrated efficacy in general tasks. However, explorations in domain-specific applications, particularly healthcare, remain insufficient. Medical question answering, with its high-stakes requirements for accuracy, ethical reasoning, and handling datasets like PubMedQA or MedQA, demands tailored D-MAS designs that integrate medical expertise without external tools like knowledge graphs or retrieval-augmented generation, yet few studies have delved deeply into this intersection.

To bridge this gap, we introduce MediHive, a novel decentralized multi-agent framework with a shared memory pool for medical reasoning. MediHive employs multiple LLM-based agents in a decentralized coordination architecture, enabling autonomous operations without a central decision-making agent. MediHive unfolds in a structured, iterative manner to foster emergent consensus, beginning with query initialization where the medical question is broadcast to the shared memory pool to provide a common starting point for all agents. This is followed by self-evolving role assignment, during which agents propose and refine specialized roles through self-reflection and review of peers' proposals to ensure diversity and relevance. Next comes the initial analysis phase, where each agent independently generates reasoning and preliminary answers based on their role, posting these contributions to the shared pool. Agents then autonomously detect divergences in the analyses. If significant disagreements arise, a conditional debate phase activates, incorporating turn-based rebuttals, defenses, and proposals to refine arguments and build toward alignment. Post-debate, the iterative shared fusion stage ensues, allowing agents to locally fuse shared information by critiquing and integrating peers' insights, thereby updating their reasonings and answers over multiple rounds until convergence criteria are met. Finally, consensus is reached through multiple rounds of deliberation, yielding a consolidated final response complete with compiled reasoning excerpts from the history. Overall, the main contributions of our work are threefold:

- We propose MediHive as a novel decentralized multi-agent framework that integrates shared memory and fusion mechanisms, offering a scalable and robust solution for LLM-based systems in medical question answering. This framework stands out by eliminating the need for a central coordinator, instead relying on autonomous agents that self-organize to handle complex queries efficiently.
- It directly targets the aforementioned challenges of centralized designs by enabling autonomous agent operations, robust fault tolerance, and clear separation of roles to prevent knowledge disarray.
- Through rigorous experiments on PubMedQA and MedQA datasets, we demonstrate the framework's validity, achieving predictive accuracies of 84.3% and 78.4% respectively, alongside improved agreement rates and efficient interactions, validating its superiority over single-LLM and centralized baselines.

We position MediHive as a coordination-protocol contribution rather than a new learning algorithm. The empirical claim of this paper is that protocol design alone—agent self-organization, conditional debate, and deterministic post-hoc aggregation—is sufficient to outperform centralized multi-agent baselines on medical question answering, without specialized training, external knowledge bases, or retrieval augmentation.

## 2 Related Work

### 2.1 LLM-based Multi-Agent

The paradigm of multi-agent systems (MAS), which involves multiple autonomous agents interacting within an environment to achieve collective goals, has been significantly revitalized by the integration of Large Language Models (LLMs). LLMs serve as the cognitive core for these agents, equipping them with advanced capabilities for planning, reasoning, and communication. Early explorations into this domain established frameworks where LLM-agents could collaborate on complex tasks. For instance, AutoGen [11] introduced a framework enabling the creation of conversational agent workflows for tasks like code generation and problem-solving, where agents with different roles (e.g., commander, writer) collaborate to fulfill user requests. Recent advancements focus on multi-agent collaboration mechanisms, enabling groups of LLM agents to coordinate on complex tasks through iterative consensus-building and role specialization [12]. AgentNet [9] proposes decentralized evolutionary coordination, where agents optimize expertise through profile evolution and peer interactions, addressing limitations in centralized systems. These developments underscore the shift toward scalable, fault-tolerant designs, though domain-specific adaptations remain a key area for exploration.

### 2.2 LLM-based Multi-Agent in Medical Domains

In healthcare, LLM-based MAS have gained traction for simulating multi-disciplinary teams (MDTs) in tasks like diagnosis, treatment planning, and question answering, where agents collaborate to handle uncertainty and interdisciplinary knowledge [13]. MedAgents [14] pioneers zero-shot medical collaboration using LLM agents in role-playing scenarios, enabling multi-round discussions for improved accuracy on datasets like MedQA without external tools. Building on this, MDAgents [15] introduces adaptive collaboration, assigning medical expertise to agents that operate independently or cooperatively based on query complexity, enhancing efficiency in decision-making. AgentHospital [16] simulates virtual hospital environments with LLM-driven doctors, nurses, and patients, modeling full care cycles and demonstrating gains in clinical reasoning through agent interactions. LLM-MedQA [17] focuses on case-based reasoning in multi-agent setups, using debates and consensus to refine answers on medical QA benchmarks. Similarly, other centralized frameworks have shown the benefits of collaboration between multiple LLMs for medical question answering [18]. Beyond question answering, such architectures have also been applied to specific clinical NLP tasks, including the automated detection of clinical problems from SOAP notes [19]. Debate protocols, as explored in multi-agent debate strategies, have been adapted for medical contexts to handle discrepancies through argumentative exchanges [20]. Despite these innovations, decentralized designs tailored to healthcare without central coordinators are underexplored, motivating our focus on scalable, robust frameworks for medical QA.

## 3 Methodology

In this section, we delineate the methodology underpinning our MediHive framework, a decentralized multi-agent system designed for medical question answering.

Centralized multi-agent systems, while effective for simple tasks, often encounter scalability bottlenecks, single points of failure, and inefficiencies in handling conflicting evidence—particularly in high-stakes domains like healthcare. To address these, MediHive adopts a decentralized coordination architecture, where agents collaborate autonomously through peer-to-peer interactions and a shared memory pool. Importantly, the shared memory pool serves solely as a passive, append-only repository for agent contributions; it performs no decision-making or coordination logic, and all reasoning, evaluation, and consensus-building are carried out autonomously by the individual agents. This design promotes resilience, adaptability, and emergent consensus without relying on a central coordinating agent or external tools, enabling robust performance on complex, interdisciplinary medical queries. Let $\mathcal{Q}$ denote the input medical query (e.g., from datasets like PubMedQA or MedQA), and let $\mathcal{A} = \{A_1, A_2, \ldots, A_N\}$ represent the set of $N$ LLM-based agents. $\mathcal{M}$ signifies the shared memory pool as an append-only, timestamped repository for all agent interactions, and $K$ indicates the maximum number of iterative fusion rounds. The framework operates under various prompting paradigms, leveraging internal collaborations to foster consensus via autonomous processes. The remainder of this section is structured according to the workflow: we begin with query initialization, followed by self-evolving role assignment, initial analysis, disagreement detection with conditional debate, iterative shared fusion, and finally consensus aggregation and output generation. Figure 1 presents an overview of the proposed framework, illustrating the end-to-end workflow and agent interactions.

### 3.1 Query Initialization and Role Assignment

The reasoning process commences with Query Initialization, where the input medical query $\mathcal{Q}$ is broadcast to all $N$ agents by being appended to the shared memory pool, $\mathcal{M}$.

This is immediately followed by the Self-Evolving Role Assignment, a single-round "warm-up" phase designed to establish specialized, complementary roles without a central coordinator. This phase unfolds in two steps. First, each agent $A_i \in \mathcal{A}$ independently analyzes $\mathcal{Q}$ and generates an initial role proposal, $R_{i,0}$. This proposal includes not only the intended specialization (e.g., "Pulmonologist") but also a brief reasoning for selecting that role based on the query. This proposal is then appended to $\mathcal{M}$. Second, after all initial proposals are posted, each agent $A_i$ reads the full set of proposals $\{R_{j,0}\}_{j=1}^{N}$ from $\mathcal{M}$. The agent then performs a self-reflection step to update its role from $R_{i,0}$ to a final, refined role $R_i$. This refinement, which also includes a rationale for the update, is guided by a prompt instructing the agent to optimize for three internal metrics: Clarity (a well-defined specialty), Differentiation (low semantic overlap with peers' roles), and Alignment (high relevance to $\mathcal{Q}$). Each agent posts its final role $R_i$ to $\mathcal{M}$, ensuring the set of agents $\mathcal{A}$ adopts a diverse and query-relevant distribution of expertise before the main analysis begins. An example of the output for this phase is shown in Fig. 2.

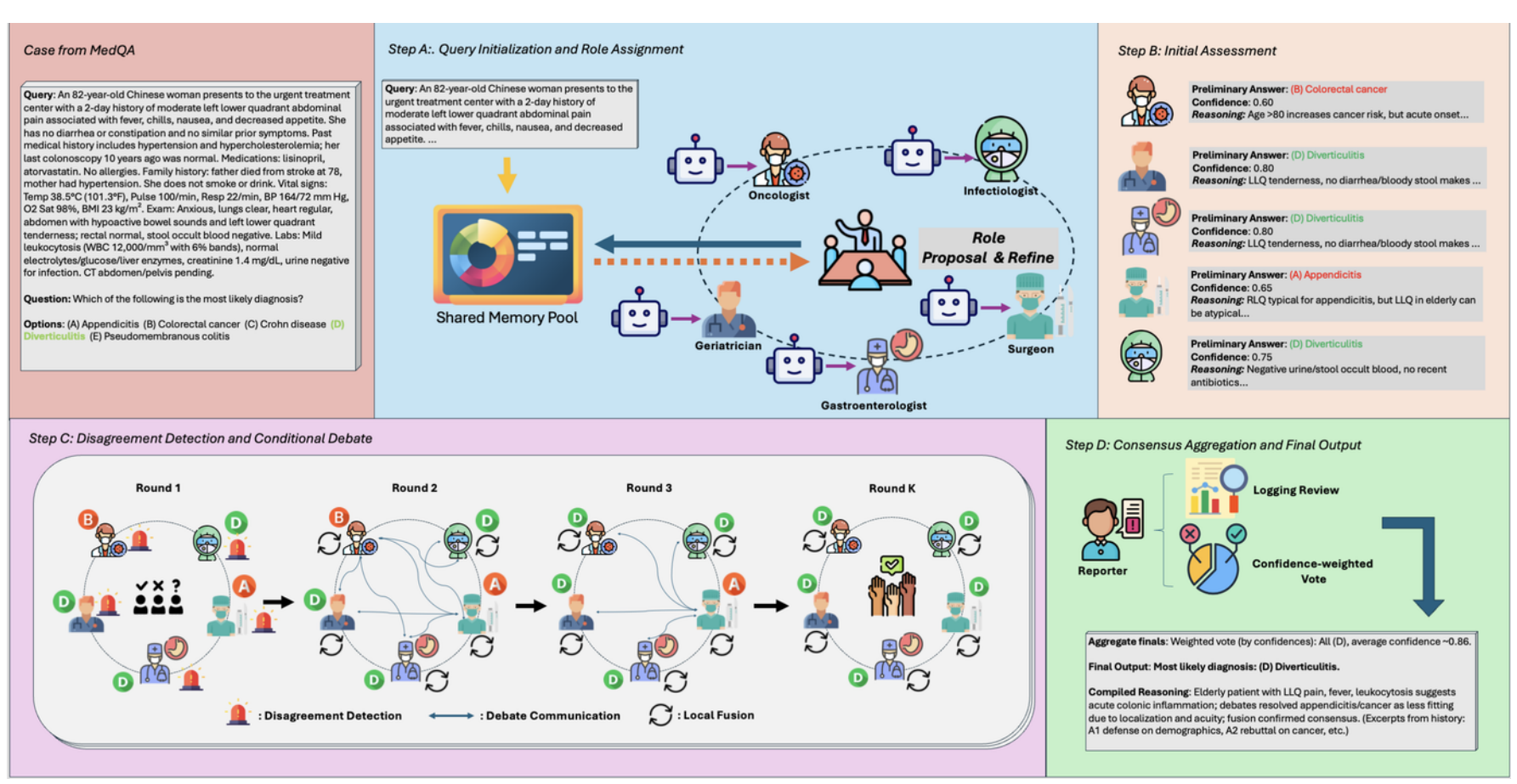

**Fig. 1** Overview of the MediHive framework's decentralized workflow for medical question answering, illustrated with a sample query from the MedQA dataset. The process unfolds in four key steps: (**A**) Query initialization and autonomous role assignment among LLM agents via the shared memory pool; (**B**) Initial assessments by specialized agents, including preliminary diagnoses with confidence scores and reasoning; (**C**) Disagreement detection triggering conditional multi-round debates, followed by local fusion of insights; and (**D**) Consensus aggregation through confidence-weighted voting, culminating in the final compiled reasoning and output

## 3.2 Analysis, Debate, and Iterative Fusion

Following the role assignment, the core reasoning process of MediHive unfolds. This multi-stage process is designed to generate independent initial insights, identify and resolve conflicts through a structured debate, and finally converge on a robust, group-vetted consensus through iterative refinement. Figure 3 contrasts centralized and decentralized designs, highlighting MediHive's key advantages in coordination, scalability, and fault tolerance.

### 3.2.1 Initial Analysis and Confidence Assessment

In the initial formal reasoning round, each agent $A_i \in \mathcal{A}$ begins its work. Guided by its finalized role $R_i$, each agent independently analyzes the query $\mathcal{Q}$ and the context from the role assignment phase. The agent is prompted to generate a comprehensive initial output, which includes: (1) a detailed reasoning trace (e.g., a Chain-of-Thought) explaining its diagnostic process, (2) a specific final answer $\text{Ans}_{i,1}$ (e.g., 'yes' for PubMedQA, or 'Option C' for MedQA), and (3) a self-assessed confidence score $c_{i,1} \in [0, 1]$. This initial, independent analysis serves as a critical baseline, capturing each agent's specialized perspective *before* it is influenced by its peers. This step is crucial for surfacing a diverse set of initial hypotheses and identifying the primary

**Query (Q):** An 83-year-old woman presents with a 2-day history of moderate left lower quadrant abdominal pain. Exam reveals localized tenderness with guarding. Labs: WBC 14,200/mm$^3$. What is the most likely diagnosis?
Options: (A) Appendicitis (B) Colorectal cancer (C) Colonic diverticulitis (D) Pseudomembranous colitis

**Step 1: Role Proposals** ⊠ M

**[A1] Role:** Gastroenterologist
**Reasoning:** Abdominal pain with localized tenderness and elevated WBC suggests a GI etiology requiring differential diagnosis.

**[A2] Role:** Geriatrician
**Reasoning:** Patient's advanced age (83) significantly alters the differential for abdominal pathology.

**[A3] Role:** Surgeon
**Reasoning:** LLQ tenderness with guarding and leukocytosis may indicate a surgical emergency.

**Step 2: Refined Roles** ⊠ M

**[A1] Role:** GI Specialist (Colorectal Disorders)
**Rationale:** With A2 covering age-related factors, I narrow my focus to colorectal pathology specific to the LLQ.

**[A2] Role:** Geriatric Medicine Specialist
**Rationale:** Complementary to A1 and A3. I will analyze age-related risk factors and atypical presentations.

**[A3] Role:** Surgical Consultant (Acute Abdomen)
**Rationale:** A1 handles GI diagnostics; I focus on surgical indications and the urgency of the presentation.

**Fig. 2** Illustrative output of the Role Assignment phase for the sample MedQA query from Fig. 1, showing initial proposals and peer-aware refinements posted to $\mathcal{M}$

points of disagreement. The confidence score is not a statistical probability but a self-assessed metric where the LLM is prompted to rate its conviction in its answer. This score provides a vital heuristic for agents to weigh their own and their peers' opinions in subsequent rounds. The complete initial output $(\text{Reasoning}_{i,1}, \text{Ans}_{i,1}, c_{i,1})$ is then appended to the shared memory pool $\mathcal{M}$. This description can be expressed by the following formula:

$$\text{Reasoning}_{i,1},\ \text{Ans}_{i,1},\ c_{i,1} = A_i(\mathcal{Q}, R_i, \mathcal{M}), \quad i \in \{1, \ldots, N\}. \tag{1}$$

### 3.2.2 Disagreement Detection and Conditional Debate

After all agents have posted their initial analysis, each agent $A_i$ *autonomously* executes a disagreement detection protocol—there is no central manager to check for consensus. Each agent independently reads the full set of initial answers $\{\text{Ans}_{j,1}\}_{j=1}^{N}$ from $\mathcal{M}$ and computes the current level of agreement. For tasks like PubMedQA, this is a direct tally of the 'yes'/'no'/'maybe' votes; for MedQA, it is a count of the selected options. If no single answer holds a supermajority (i.e., the agreement level falls below a predefined threshold $\tau_{agree}$, e.g., 0.8), the agents determine that significant disagreement exists, and the conditional debate phase is activated. Because every agent independently evaluates the same shared evidence, this check requires no central arbiter while ensuring that all participants reach a consistent assessment of the group's state. When agents already agree, the framework bypasses the debate and proceeds directly to the fusion stage.

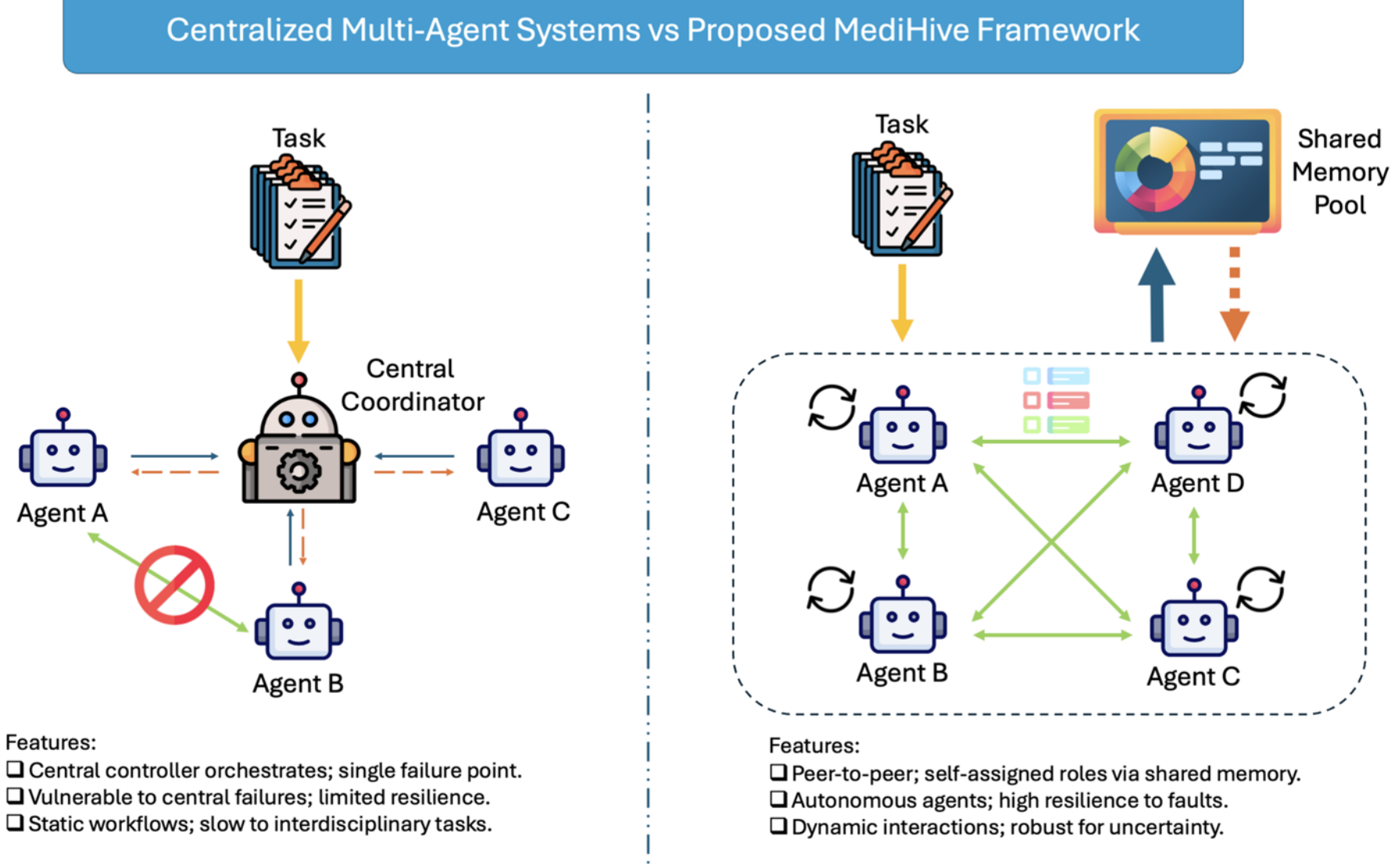

**Fig. 3** Comparison of centralized and proposed MediHive framework, highlighting coordination, resilience, and adaptability

If activated, the debate proceeds for $T_{debate}$ cycles (e.g., $T_{debate} = 2$). The purpose of this stage is adversarial: to stress-test initial hypotheses and enrich the shared evidence base before agents attempt to converge. In each cycle, all agents simultaneously contribute a structured argument to $\mathcal{M}$. Each argument takes one of three forms:

- **Rebuttal:** A targeted, evidence-based challenge to a specific peer's reasoning (e.g., "Addressing A2: Your reliance on [fact] is contradicted by..."). This exposes weaknesses in arguments and promotes critical evaluation.
- **Defense:** A reinforcement of the agent's own position, providing additional evidence or strengthening its logic in response to a peer's challenge. This ensures that well-supported positions are not prematurely abandoned.
- **Proposal:** A new, synthesized hypothesis that integrates insights from multiple peers into a revised position, bridging opposing views and seeding potential consensus for the subsequent fusion stage.

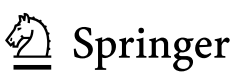

Crucially, the debate does not require agents to update their formal answers or confidence scores; its role is to generate a rich adversarial evidence base that informs the subsequent fusion stage. All debate contributions are logged to $\mathcal{M}$, creating a transparent and auditable record. The debate runs for the full $T_{debate}$ cycles to ensure thorough examination of the disagreements before the framework transitions to convergence.

### 3.2.3 Iterative Shared Fusion

Following the initial analysis (and the conditional debate, if it occurred), the framework enters the iterative shared fusion stage. While the debate is *adversarial*—designed to challenge assumptions and enrich the evidence base—the fusion stage is *integrative*: each agent synthesizes the accumulated evidence into a formal, structured position. This is the sole stage in which agents produce updated answers and confidence scores, and the sole stage in which formal convergence is measured.

The fusion process proceeds in rounds beginning at $k = 2$, with the scope of each agent's input determined by the round.

**Comprehensive Integration (Round k=2)** In the first fusion round, each agent $A_i$ reads the entire interaction history stored in $\mathcal{M}$: the initial analyses from all peers ($k = 1$) and, if a debate was triggered, its full argumentative log. The agent then executes a structured integration process: (1) **Critique** the strengths and weaknesses of peer arguments, (2) **Integrate** the most compelling evidence into its own reasoning, and (3) **Revise** its initial stance accordingly. This comprehensive review is essential because it is the first point at which agents formally update their positions in light of all accumulated evidence. This yields the agent's first fusion-based entry: $(\text{Reasoning}_{i,2}, \text{Ans}_{i,2}, c_{i,2})$.

**Incremental Refinement (Rounds $k > 2$)** In subsequent rounds (up to the maximum *K*), each agent reads the full set of peer outputs from the immediately preceding round, $\{(\text{Reasoning}_{j,k-1}, \text{Ans}_{j,k-1}, c_{j,k-1})\}_{j=1}^{N}$, rather than the entire history. This narrowing of scope is justified because the comprehensive review at $k$=2 has already incorporated the debate's insights; further rounds serve to refine positions in response to peers' evolving stances. Each agent assesses the current group consensus, critiques peers' latest reasoning, and updates its own output, producing $(\text{Reasoning}_{i,k}, \text{Ans}_{i,k}, c_{i,k})$.

The fusion loop terminates when the agreement level reaches or exceeds $\tau_{agree}$ for two consecutive rounds, ensuring that convergence is stable rather than transient, or when the maximum round limit *K* is reached. The complete end-to-end workflow of MediHive is summarized in Algorithm 1.

**Algorithm 1** The MediHive decentralized multi-agent reasoning workflow.

1: **procedure** MEDIHIVE($\mathcal{Q}$, $\mathcal{A}$, $K$, $T_{debate}$, $\tau_{agree}$)
2: $\mathcal{M} \leftarrow \{\mathcal{Q}\}$ ▷ Initialize shared memory pool with query
▷ **Phase 1: Role Assignment**
3: **for** each $A_i \in \mathcal{A}$ **do** ▷ Step 1: Independent role proposals
4: $R_{i,0} \leftarrow A_i$.ProposeRole($\mathcal{Q}$); Append($\mathcal{M}$, $R_{i,0}$)
5: **end for**
6: **for** each $A_i \in \mathcal{A}$ **do** ▷ Step 2: Peer-aware refinement
7: $R_i \leftarrow A_i$.RefineRole($\mathcal{M}$); Append($\mathcal{M}$, $R_i$)
8: **end for**
▷ **Phase 2: Initial Analysis** ($k$=1)
9: **for** each $A_i \in \mathcal{A}$ **do**
10: ($\text{Reas}_{i,1}$, $\text{Ans}_{i,1}$, $c_{i,1}$) $\leftarrow A_i$.Analyze($\mathcal{Q}$, $R_i$, $\mathcal{M}$); Append($\mathcal{M}$, $\cdot$)
11: **end for**
▷ **Phase 3: Disagreement Detection and Conditional Debate**
12: **if** $\max(\text{Tally}(\{\text{Ans}_{i,1}\}))/N < \tau_{agree}$ **then**
13: **for** $t \leftarrow 1$ **to** $T_{debate}$ **do** ▷ Debate: Rebuttal, Defend, or Propose
14: **for** each $A_i \in \mathcal{A}$ **do**
15: Append($\mathcal{M}$, $A_i$.Debate($\mathcal{M}$))
16: **end for**
17: **end for**
18: **end if**
▷ **Phase 4: Iterative Shared Fusion** ($k$=2 to $K$)
19: prevAgrmt $\leftarrow 0$
20: **for** $k \leftarrow 2$ **to** $K$ **do**
21: $\text{ctx} \leftarrow \begin{cases} \mathcal{M}.\text{readAll}() & \text{if } k = 2 \\ \{(\text{Reas}_{j,k-1}, \text{Ans}_{j,k-1}, c_{j,k-1})\}_{j=1}^{N} & \text{if } k > 2 \end{cases}$
22: **for** each $A_i \in \mathcal{A}$ **do**
23: ($\text{Reas}_{i,k}$, $\text{Ans}_{i,k}$, $c_{i,k}$) $\leftarrow A_i$.Fuse(ctx); Append($\mathcal{M}$, $\cdot$)
24: **end for**
25: currAgrmt $\leftarrow \max(\text{Tally}(\{\text{Ans}_{i,k}\}))/N$
26: **if** currAgrmt $\geq \tau_{agree}$ **and** prevAgrmt $\geq \tau_{agree}$ **then break** ▷ Stable consensus
27: **end if**
28: prevAgrmt $\leftarrow$ currAgrmt
29: **end for**
▷ **Phase 5: Final Output (Reporter)**
30: $k^* \leftarrow$ last round
31: $a^* \leftarrow \begin{cases} \text{MajorityAnswer}(\{\text{Ans}_{i,k^*}\}) & \text{if agreement} \geq \tau_{agree} \\ \arg\max_a \sum_i c_{i,k^*} \cdot \mathbf{1}(\text{Ans}_{i,k^*}{=}a) & \text{otherwise} \end{cases}$
32: **return** ($a^*$, SynthesizeTrace($\mathcal{M}$))
33: **end procedure**

### 3.3 Final Synthesis by the Reporter

Upon the termination of the iterative fusion loop, the framework transitions from decentralized collaborative reasoning to a final synthesis stage. This stage is handled by a specialized Reporter, whose role is strictly limited to aggregating and summarizing the independently generated outputs of the reasoning agents, rather than coordinating or influencing their decision processes.

The Reporter operates in a post-hoc manner and applies an adaptive aggregation strategy depending on how the fusion loop concludes. Specifically, the Reporter first determines the termination condition of the loop:

- **If a strong consensus was reached** (i.e., the loop terminated because agreement surpassed the threshold $\tau_{agree}$), the Reporter's role is confirmatory. It identifies the answer supported by the supermajority of agents and designates it as the final answer, $a^*$. In this case, no additional decision logic is introduced, as collective agreement has already emerged from decentralized agent reasoning.
- **If no consensus was reached** (i.e., the loop terminated upon reaching the maximum round limit $K$), the Reporter applies a predefined resolution mechanism in the form of a **confidence-weighted vote**. Importantly, this procedure does not alter agent reasoning or retroactively enforce agreement, but serves solely as a deterministic aggregation rule to produce a single output from divergent agent conclusions. The process consists of two steps:
  First, the Reporter computes a total confidence score, $S(a)$, for each candidate answer $a$ in the set of all possible answers $\mathcal{V}$, by summing the confidence scores ($c_{i,k}$) of agents whose final answer ($a_{i,k}$) matches $a$:

$$S(a) = \sum_{i=1}^{N} c_{i,k} \cdot 1(a_{i,k} = a)$$

  Second, the final answer is selected as:

$$a^* = \arg\max_{a \in \mathcal{V}} S(a)$$

Beyond answer aggregation, the Reporter also provides an explanatory summary of the reasoning process. It constructs a reasoning trace by parsing the shared memory $\mathcal{M}$, which stores the historical outputs generated by agents during the fusion loop. This trace highlights salient elements of the interaction history, including initial role assignments, key debate exchanges, and representative reasoning steps.

The final output consists of the selected answer $a^*$ together with its supporting reasoning trace. This design supports transparency by exposing the intermediate reasoning artifacts generated during decentralized agent interaction, thereby facilitating post-hoc inspection and analysis without introducing centralized control into the reasoning process.

## 4 Experiments

### 4.1 Dataset

We evaluate our proposed framework on two widely used benchmarks for medical question answering: PubMedQA [21] and MedQA [22], which test the system's ability to handle biomedical reasoning and clinical knowledge. PubMedQA is a dataset designed for biomedical research question answering, comprising questions paired with PubMed abstracts that require yes/no/maybe answers along with supporting reasoning. It emphasizes the need for quantitative and evidence-based inference over scientific texts. We focus on the test set for evaluation, assessing performance in a closed-book setting where agents rely solely on internal knowledge and collaboration. MedQA is a large-scale open-domain question answering dataset derived from United States Medical Licensing Examination (USMLE) questions, featuring multiple-choice formats with four or five options. This dataset challenges models on comprehensive medical knowledge, including diagnosis, treatment, and ethical considerations, making it ideal for validating multi-agent reasoning in high-stakes scenarios. Our experiments utilize the test split to measure accuracy in zero-shot collaborative settings.

### 4.2 Main Results

In this section, we present a comprehensive evaluation of our proposed framework, MediHive, focusing on its zero-shot accuracy and F1-score advantages on the MedQA and PubMedQA datasets. All experiments were conducted using Llama-3.1-70B-Instruct as the base large language model. All baselines, including multi-agent methods originally evaluated with different models, were reimplemented under this same backbone to ensure a fair comparison across all configurations. The results, summarized in Table 1, benchmark our method against two primary categories of baselines. The **Single-Agent** configurations evaluate the performance of the base LLM under various prompting strategies: standard zero-shot, zero-shot with Chain-of-Thought (w/ CoT) to elicit step-by-step reasoning, and an enhanced version that adds Self-Consistency (w/ CoT + SC) by sampling multiple reasoning paths. These baselines establish a performance ceiling for non-collaborative approaches, topping out at a 74.1% average accuracy. The **Multi-Agent** baselines demonstrate the inherent benefits of collaborative reasoning, setting a higher competitive standard. These include a centralized multi-agent system, in which a single coordinator agent assigns specialist roles to $N=5$ agents, moderates their multi-round discussion, and synthesizes the final answer—a design consistent with standard centralized configurations in the literature [1, 15]—as well as state-of-the-art frameworks: MedAgents [14], which uses role-based collaborative consultation with centralized report synthesis, and Multiagent Debate [23], which improves factuality through iterative multi-agent debate rounds.

**Table 1** Main results on accuracy and F1-score across MedQA and PubMedQA datasets

| Method | MedQA | | PubMedQA | | Avg. |
|---|---|---|---|---|---|
| | Accuracy (%) | F1-score (%) | Accuracy (%) | F1-score (%) | Accuracy (%) |
| **Single-Agent** | | | | | |
| **zero-shot setting* | | | | | |
| Zero-shot | 71.5 | 66.8 | 69.3 | 67.8 | 70.4 |
| Zero-shot w/ CoT | 73.2 | 70.5 | 70.7 | 69.2 | 72.0 |
| Zero-shot w/ CoT + SC | 73.8 | 70.9 | 72.1 | 70.6 | 73.0 |
| **few-shot setting* | | | | | |
| Few-shot | 72.8 | 71.0 | 71.5 | 69.9 | 72.2 |
| Few-shot w/ CoT | 74.6 | 73.1 | 72.3 | 70.7 | 73.5 |
| Few-shot w/ CoT + SC | 74.4 | 72.9 | 73.8 | 72.1 | 74.1 |
| **Multi-Agent** | | | | | |
| Centralized Multi-Agent System | 77.8 | 76.1 | 75.3 | 74.1 | 76.6 |
| MedAgents [14] | 83.7 | 82.1 | 76.8 | 75.1 | 80.3 |
| Multiagent Debate [23] | 80.4 | 78.3 | 78.2 | 76.4 | 79.3 |
| MediHive (Ours) | **84.3** | **82.5** | **78.4** | **76.8** | **81.4** |

Our framework, MediHive, operates in the same zero-shot setting, leveraging its decentralized architecture with five agents (N = 5) to achieve state-of-the-art performance. As shown in the results, MediHive attains an overall average accuracy of 81.4%, driven by strong individual performances of 84.3% on MedQA and 78.4% on PubMedQA. This represents a substantial improvement of 7.3 percentage points over the strongest single-agent method. More importantly, MediHive outperforms the leading multi-agent baselines. For instance, while MedAgents achieves a commendable 80.3% average accuracy, MediHive surpasses it by 1.1 points, with a particularly strong showing on the complex MedQA dataset. This superior performance validates the efficacy of our framework's unique components—such as self-evolving roles, conditional debate, and efficient iterative fusion—in fostering a more robust and accurate consensus for complex medical reasoning.

### 4.3 Ablation Study

To evaluate the contributions of key components in the MediHive framework, we conducted an ablation study on the MedQA and PubMedQA datasets, with results summarized in Table 2. The full MediHive framework achieves accuracies of 84.3% on MedQA and 78.4% on PubMedQA, serving as the baseline. We target three modular components that can be cleanly removed without altering the overall pipeline structure: Chain-of-Thought (CoT) reasoning, Self-Evolving Role Assignment, and Confidence-Weighted Voting. Removing CoT reasoning, which guides agents to produce step-by-step logical inferences, results in a noticeable performance drop to 78.0% on MedQA and 73.0% on PubMedQA. This decline underscores the importance of structured reasoning for navigating the complex, evidence-based requirements of medical question answering, particularly in datasets demanding nuanced interpretation. Disabling Self-Evolving Role Assignment, where agents dynamically specialize based on query context and peer proposals, reduces accuracies to 81.5% on MedQA and 75.9% on PubMedQA. This indicates that adaptive role specialization is critical for ensuring diverse expertise, preventing redundant analyses, and enhancing collaborative accuracy. Omitting Confidence-Weighted Voting, which prioritizes high-confidence outputs during final aggregation, yields accuracies of 82.4% on MedQA and 76.6% on PubMedQA. The smaller drop suggests that while weighted voting refines consensus by leveraging agent certainty, majority voting still captures collective decisions, albeit with reduced precision. These results validate that each component—CoT, role assignment, and weighted voting—plays a significant role in MediHive's effectiveness, with CoT and role assignment being particularly crucial for high-stakes medical tasks. We note that the conditional debate and iterative fusion mechanisms are structural components of the pipeline whose removal would fundamentally alter the system's architecture rather than isolate a single variable; their collective contribution is instead reflected in MediHive's consistent advantage over the centralized baseline, which lacks both mechanisms.

**Table 2** Ablation study results

| Method | MedQA(%) | PubMedQA(%) |
|---|---|---|
| MediHive | 84.3 | 78.4 |
| w/o CoT | 78.0 | 73.0 |
| w/o Self-Evolving Role Assignment | 81.5 | 75.9 |
| w/o Confidence-Weighted Voting | 82.4 | 76.6 |

### 4.4 Impact of the Number of Agents

As our MediHive framework is predicated on the collaboration between multiple autonomous agents, we explored how the number of agents ($N$) influences the system's overall performance. We varied the number of agents in the system, testing configurations of $N \in \{3, 5, 7\}$, while holding all other parameters constant. The results of this analysis on both the MedQA and PubMedQA datasets are illustrated in Fig. 4.

Our key observation is that performance is not monotonically increasing with the number of agents. Instead, we find a consistent peak for both datasets, with the optimal accuracy being achieved with a configuration of five agents. The performance degradation with fewer agents ($N = 3$) suggests that a smaller group may lack the necessary diversity of self-assigned roles to comprehensively address the query. Conversely, increasing the group to seven agents also leads to a decline in accuracy. We hypothesize this is due to an increase in conversational noise and the greater difficulty in reaching a stable consensus, potentially caused by role redundancy. This finding suggests that five agents strike an optimal balance between expert diversity and efficient, focused collaboration. Based on this analysis, we set $N = 5$ as the default configuration for all other experiments in this paper.

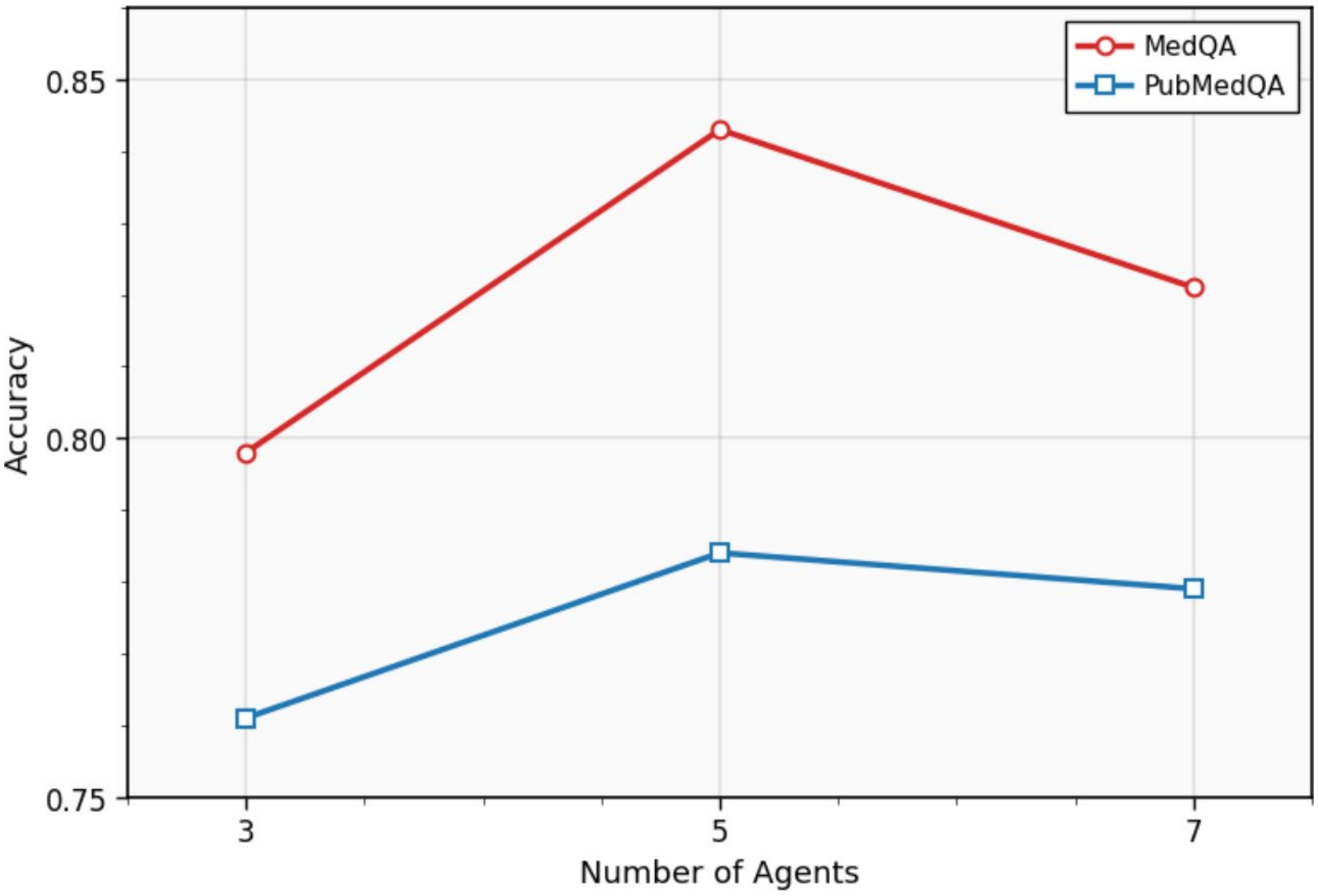

**Fig. 4** Performance on MedQA and PubMedQA datasets as a function of the number of collaborating agents ($N$). The optimal accuracy for both datasets is achieved with $N = 5$

## 5 Limitations and Future Work

Despite its promising results, our framework has several limitations that present clear opportunities for future research.

First, the framework's effectiveness has been demonstrated on controlled medical question answering benchmarks. While such datasets are commonly used to study medical reasoning, they do not fully capture the complexity, ambiguity, and contextual richness of real-world clinical workflows. As a result, MediHive is not intended for direct clinical decision-making, and real-world clinical validation remains an important future direction.

Second, our evaluation primarily focuses on mean performance trends under controlled benchmark settings, without conducting statistical significance testing, confidence interval analysis, or a detailed assessment of computational efficiency. While this choice allows us to concentrate on system-level coordination behavior, a more comprehensive evaluation framework that integrates statistical rigor with efficiency and scalability analysis will be an important direction for future work.

Third, our evaluation is restricted to LLM-based baselines in a closed-book setting and does not include comparisons with retrieval-augmented hybrid pipelines or non-LLM clinical decision support models. Such comparisons would help isolate the contribution of decentralized coordination from that of LLM capacity alone, and would clarify the regime in which decentralized LLM coordination is preferable to retrieval-based or symbolic approaches. Integrating a Retrieval-Augmented Generation (RAG) pipeline and benchmarking against these alternative architectures is a key direction for future work, both to ground agent reasoning in current medical literature and to reduce hallucinations.

Finally, although our framework achieves strong aggregate accuracy, qualitative inspection of incorrect predictions reveals several recurring failure modes that warrant systematic study. We observed cases of *collective high-confidence error*, in which all agents converge on the same incorrect answer with uniformly high confidence—typically on queries involving rare conditions or uncommon presentations that lie outside the backbone model's well-represented distribution. Performing a systematic safety or clinical risk assessment—adversarial robustness testing, and safety-aware evaluation metrics—will be critical for any future translation toward clinical application.

## 6 Conclusion

In this work, we introduced MediHive, a decentralized multi-agent framework designed to support complex medical reasoning through structured coordination. Rather than relying on a central coordinator, MediHive enables agents to operate with autonomous reasoning roles, adapt dynamically to the input query, and reach consensus via conditional debate and iterative information fusion. The decentralization is preserved throughout the reasoning loop: the shared memory pool serves as a passive, append-only repository with no coordination logic, and the final aggregation stage is a post-hoc deterministic rule rather than a central decision-maker that influences agent reasoning.

Experimental results on the MedQA and PubMedQA benchmarks demonstrate that MediHive consistently outperforms single-agent approaches and competitive multi-agent baselines under the same model backbone, highlighting the effectiveness of decentralized coordination in medical question answering. In addition, ablation studies confirm that each architectural component contributes meaningfully to the overall system performance.

Together, these findings suggest that decentralized role-based coordination offers a promising direction for improving robustness and reasoning quality in LLM-based medical QA systems, and provide a foundation for future research on scalable, reliable multi-agent reasoning frameworks in healthcare contexts.

**Acknowledgements** This work was supported in part by the National Science Foundation under the Grants IIS-1741306 and IIS-2235548, and by the Department of Defense under the Grant DoD W91XWH-05-1-023. This material is based upon work supported by (while serving at) the National Science Foundation. Any opinions, findings, conclusions, or recommendations expressed in this material are those of the author(s) and do not necessarily reflect the views of the National Science Foundation.

**Author Contributions** X.W. conceived the MediHive framework, designed the methodology, implemented the system, conducted all experiments and analyses, prepared the figures and tables, and drafted the manuscript. C.C.Y. supervised the project, provided guidance on the research direction and experimental design, secured funding and computational resources, and critically reviewed and revised the manuscript. Both authors read and approved the final version of the manuscript.

**Funding** This work was supported in part by the National Science Foundation under Grants IIS-1741306 and IIS-2235548, and by the Department of Defense under Grant DoD W91XWH-05-1-023. This material is based upon work supported by (while serving at) the National Science Foundation.

**Data Availability** No datasets were generated or analysed during the current study.

**Materials Availability** Not applicable.

**Code Availability** The source code implementing the MediHive framework, including agent prompt templates, the shared memory pool, and evaluation scripts, will be made publicly available on GitHub upon acceptance of this manuscript. The repository URL will be provided in the final published version. In the interim, the code is available from the corresponding author upon reasonable request.

## Declarations

**Ethics approval and consent to participate** Not applicable. This study did not involve human participants, human-derived data, or animal subjects. All experiments were conducted on publicly available, de-identified benchmark datasets.

**Consent for publication** Not applicable.

**Competing interests** Christopher C. Yang serves as Editor-in-Chief of the Journal of Healthcare Informatics Research. He recused himself from all editorial decisions, reviewer assignment, and access to confidential review materials for this manuscript, which has been handled independently by another member of the Editorial Board. The authors declare no other competing financial or non-financial interests